\documentclass[11pt]{article}
\usepackage[final]{acl}

\usepackage{times}
\usepackage{latexsym}
\usepackage[T1]{fontenc}
\usepackage[utf8]{inputenc}
\usepackage{microtype}
\usepackage{inconsolata}
\usepackage{graphicx}
\usepackage[english]{babel}
\usepackage{amsmath}
\usepackage{amssymb}
\usepackage{booktabs}
\usepackage{tabularx}
\usepackage{multirow}
\usepackage{listings}
\usepackage{tikz}
\usetikzlibrary{arrows.meta, positioning, shapes.multipart}

\usepackage{enumitem}
\setlist{itemsep=2pt, topsep=2pt, parsep=0pt, partopsep=0pt}

\title{Multilingual TinyStories: A Synthetic Combinatorial Corpus of Indic Children's Stories for Training Small Language Models}

\author{
  Deepon Halder \\
  AI4Bharat, IIEST Shibpur \\
  \And
  Angira Mukherjee \\
  IIEST Shibpur \\
}

\begin{document}
\maketitle

\begin{abstract}
The development of robust language models for low-resource languages is frequently bottlenecked by the scarcity of high-quality, coherent, and domain-appropriate training corpora. In this paper, we introduce the Multilingual TinyStories dataset, a large-scale, synthetically generated collection of children's stories encompassing 17 Indian languages. Designed specifically for the training and evaluation of Small Language Models (SLMs), the corpus provides simple, narrative-driven text strictly localized to native scripts. We detail our hybrid curation pipeline, which leverages the Sarvam-M language model and a novel combinatorial prompt engineering framework for native generation, coupled with the Google Translate API for large-scale cross-lingual expansion. Through strict programmatic filtering, we compiled 132,942 stories and over 93.9 million tokens in our release, serving as a foundational resource for multilingual language modeling and transfer learning in the Indic linguistic sphere.
\end{abstract}

\section{Introduction}
Recent advancements in Natural Language Processing (NLP) have been largely driven by the exponential scaling of Large Language Models (LLMs). However, the zero-shot capabilities of these models remain disproportionately skewed toward high-resource languages, predominantly English. For the vast linguistic diversity of the Indian subcontinent, the lack of extensive, high-quality, and structurally simple corpora poses a significant barrier to democratizing artificial intelligence capabilities.

Training models from scratch or aligning existing models to low-resource languages requires data that is not merely voluminous, but also clean and linguistically coherent. Inspired by the original TinyStories dataset \citep{eldan2023tinystories}, which demonstrated that SLMs can learn syntax, semantics, and reasoning from constrained, simple vocabularies, we propose a multilingual extension tailored to Indic languages.

The \textbf{Multilingual TinyStories} dataset provides a comprehensive, synthetic corpus of children's stories. By focusing on simple narratives, calibrated to the reading comprehension of a five-year-old child, the dataset intentionally restricts vocabulary complexity. This allows small-parameter models to effectively capture core linguistic structures without being overwhelmed by domain-specific jargon or web-scraped noise. To avoid redundancy with existing efforts, we explicitly exclude languages such as Bengali, Marathi, and Hindi, focusing instead on underrepresented demographics.

In summary, the primary contributions of our work are as follows:
\begin{itemize}
    \item \textbf{A Novel Multilingual Corpus:} We introduce the Multilingual TinyStories dataset, a large-scale, open-source synthetic corpus comprising over 132,000 children's stories and approximately 93.9 million tokens across 17 underrepresented Indic languages.
    \item \textbf{Tailored Resource for SLMs:} We provide a clean, structurally simple, and highly constrained dataset strictly localized to native scripts, addressing the critical lack of foundational pre-training and alignment resources for Small Language Models in the Indic linguistic sphere.
\end{itemize}
\begin{figure*}
    \centering
    \includegraphics[width=0.873\linewidth]{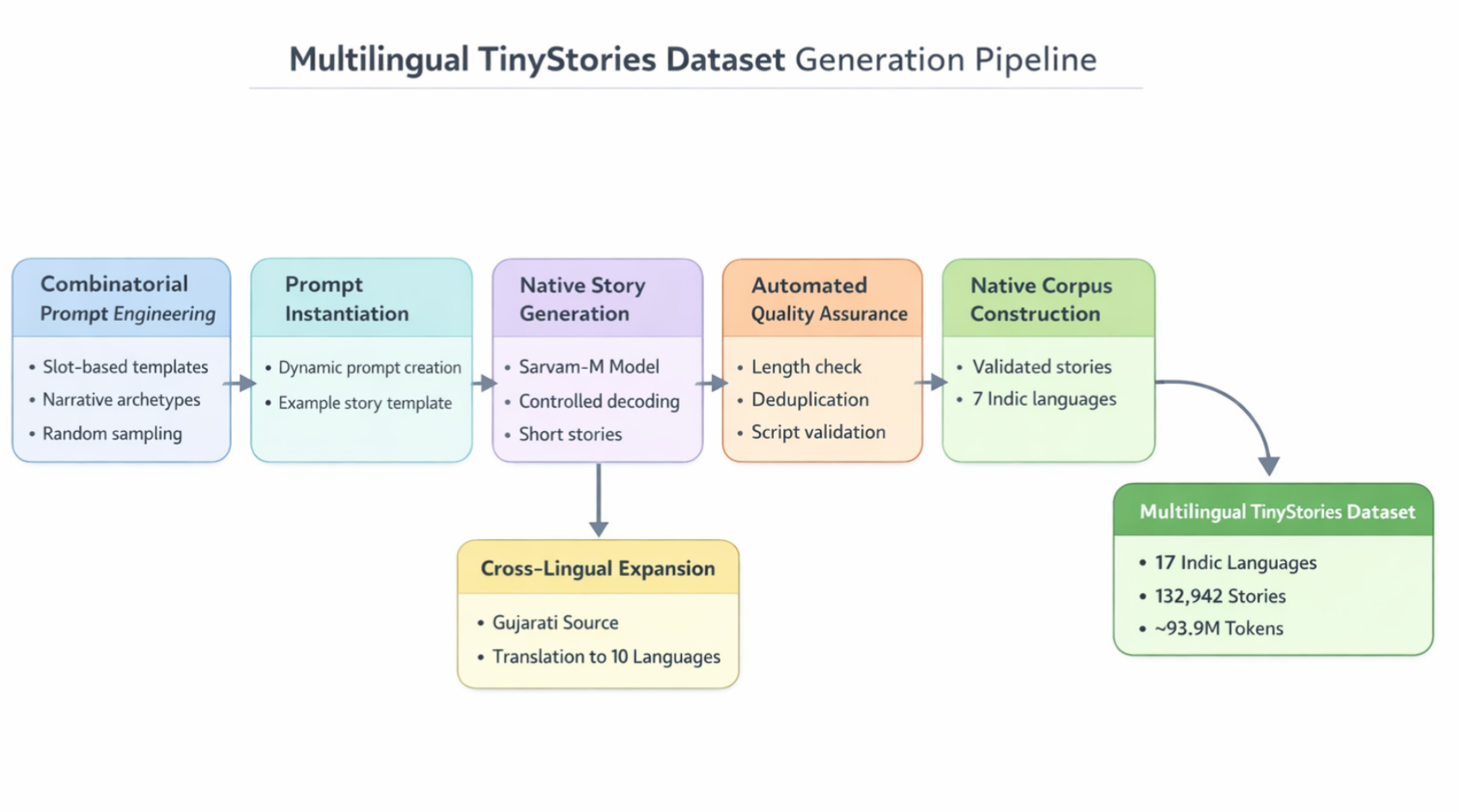}
    \caption{Pipeline for generating the Multilingual TinyStories dataset. Slot based prompts are instantiated and used to generate native stories with the Sarvam-M model, followed by automated filtering and corpus construction across seven Indic languages. The dataset is then expanded via translation to additional languages, producing 132,942 stories in 17 Indic languages (~93.9M tokens).}
    \label{fig:placeholder}
\end{figure*}
\section{Related Work}
Our work intersects with several active areas of research in natural language processing: synthetic data generation, small language models, and low-resource multilingual corpora.

\textbf{Synthetic Data for SLMs:} The efficacy of using highly constrained synthetic text to train capable language models was popularized by the original TinyStories dataset \citep{eldan2023tinystories}. They demonstrated that a model with merely 10 million parameters could produce fluent, consistent narratives when trained on synthetically generated children's stories. This paradigm was further validated by the "Textbooks Are All You Need" approach \citep{gunasekar2023textbooks}, proving that data quality and structural clarity can effectively substitute for raw parameter scale.

\textbf{Indic Language Corpora:} Historically, NLP research in Indian languages has relied on massive, web-crawled datasets such as IndicCorp \citep{kakwani2020indicnlpsuite} and parallel translation corpora like Samanantar \citep{ramesh2022samanantar}. While these resources are invaluable for large-scale pre-training, they are inherently noisy, formally complex, and unstructured, making them suboptimal for the initial syntactical alignment of small-parameter models. 

\textbf{Multilingual Synthetic Corpora:} Recent efforts have begun adapting the TinyStories methodology to non-English languages. Notably, the Regional TinyStories project by Vizuara \citep{vizuara2025regional} initiated this transition by focusing on high-resource Indian languages. Our Multilingual TinyStories corpus directly complements this literature by expanding the narrative methodology to 17 heavily underrepresented Indic languages via a hybrid generation and translation pipeline.

\section{Methodology}
To systematically generate millions of high-quality tokens across varied linguistic distributions, we designed a scalable Python pipeline. The methodology employs a hybrid approach: native synthetic generation via the Sarvam-M language model \citep{sarvam2024sarvamm} for a core set of languages, followed by machine translation to project the dataset into further low-resource domains.

\subsection{Combinatorial Prompt Engineering}
For the foundational generated corpus, we avoided static prompts to prevent mode collapse. Instead, we developed a highly dynamic, slot-based prompt template system capable of generating millions of unique prompt permutations. 

Our framework utilizes over 15 distinct narrative archetypes (e.g., \textit{Adventure}, \textit{Moral}, \textit{Mystery}, \textit{Friendship}). The full set of prompt templates used in the generation pipeline is provided in Appendix~\ref{app:prompt_templates}. Each template contains variable slots dynamically populated during runtime by uniformly sampling from extensively curated conceptual pools:
\begin{itemize}
    \item \textbf{Characters ($N \ge 50$):} Archetypal entities (e.g., \textit{boy, girl, rabbit, robot, princess}).
    \item \textbf{Settings ($N \ge 40$):} Spatial contexts (e.g., \textit{forest, village, space station, desert}).
    \item \textbf{Objects/Problems ($N \ge 50$):} Narrative catalysts (e.g., \textit{lost lantern, hidden treasure}).
    \item \textbf{Themes ($N \ge 30$):} Underlying didactic lessons (e.g., \textit{bravery, curiosity, honesty}).
\end{itemize}
The complete vocabulary pools used to populate these slots are listed in Appendix~\ref{app:slot_vocab}.

An example instantiated prompt takes the following form:
\begin{quote}
\small
\textit{Write a short children's story in \{Gujarati\}. The story should include a \{curious rabbit\}, a \{forest\}, and a problem involving a \{lost lantern\}. The story should teach something about \{bravery\}. Keep the language simple for a 5-year-old. Use exactly 5 to 8 sentences.}
\end{quote}
This stochastic slot-filling strategy naturally forces the Sarvam-M model into disparate regions of its latent space, ensuring high semantic variance. A quantitative estimate of the theoretical prompt combinatorial space is presented in Appendix~\ref{app:prompt_space}.

\subsection{Decoding Strategy and Generation Constraints}
Stochastic decoding was enforced with a \texttt{temperature} of 0.9 and a nucleus sampling threshold (\texttt{top\_p}) of 0.95. Furthermore, we imposed a strict length constraint via \texttt{max\_new\_tokens = 400}, driving the model to construct concise stories (averaging 80 to 200 tokens) with a clear climax within a 5 to 8 sentence limit.
\begin{table}[t]
\centering
\caption{Lexical diversity statistics across all language splits.}
\label{tab:diversity}
\resizebox{\columnwidth}{!}{
\begin{tabular}{lcccc}
\toprule
Lang & Dist-1 & Dist-2 & Dist-3 & AvgSim \\
\midrule
as  & 0.0763 & 0.5057 & 0.8125 & 0.1486 \\
doi & 0.0462 & 0.3681 & 0.7099 & 0.2812 \\
gom & 0.0913 & 0.5518 & 0.8382 & 0.1702 \\
gu  & 0.0878 & 0.5383 & 0.8500 & 0.1917 \\
kn  & 0.1427 & 0.6660 & 0.9227 & 0.1112 \\
mai & 0.0464 & 0.3861 & 0.7311 & 0.2166 \\
ml  & 0.1711 & 0.6954 & 0.9134 & 0.1449 \\
mni & 0.0737 & 0.4434 & 0.7729 & 0.2022 \\
ne  & 0.0859 & 0.5405 & 0.8325 & 0.1746 \\
or  & 0.0919 & 0.5155 & 0.7900 & 0.1475 \\
pa  & 0.0469 & 0.3334 & 0.6788 & 0.2710 \\
sa  & 0.1063 & 0.5420 & 0.8157 & 0.1909 \\
sat & 0.0285 & 0.2556 & 0.5728 & 0.3819 \\
sd  & 0.0434 & 0.3569 & 0.6937 & 0.3053 \\
ta  & 0.1360 & 0.6563 & 0.9132 & 0.1470 \\
te  & 0.1153 & 0.5894 & 0.8787 & 0.1633 \\
ur  & 0.0418 & 0.3340 & 0.6737 & 0.3487 \\
\bottomrule
\end{tabular}}
\end{table}

\subsection{Cross-Lingual Expansion via Machine Translation}
While the primary corpus of seven languages was natively generated using the Sarvam-M model, scaling to the remaining 10 heavily underrepresented languages required a machine translation pipeline. To achieve this, the natively generated Gujarati split, chosen for its high structural fidelity and rich, clean vocabulary, served as the source corpus. 

We utilized the Google Translate API to programmatically translate the validated Gujarati dataset into the 10 target languages (Assamese, Dogri, GOM, Maithili, Manipuri, Nepali, Sanskrit, Santali, Sindhi, and Urdu). This two-staged approach ensures that the thematic and structural diversity established by the combinatorial prompt framework is accurately preserved across all 17 language splits. Representative examples from multiple language splits are provided in Appendix~\ref{app:dataset_examples}.

\section{Dataset Statistics}
The dataset is structured into distinct linguistic splits, formatted as \texttt{JSONL} records. Each entry maintains a unique identifier (e.g., \texttt{gu\_00001}). 

As detailed in Table \ref{tab:dataset_stats}, the core natively generated release encapsulates 7 languages, which was then expanded to an additional 10 languages via our translation pipeline. This combined effort yields a robust final dataset containing a total of 132,942 stories and 93,909,863 tokens. Gujarati was selected as the pivot language due to its high generation quality and minimal script contamination during preliminary experiments.
\subsection{Script and Language Normalization}
Generated stories occasionally contained Latin characters or embedded English tokens, which are undesirable for training monolingual language models. We implemented a script-normalization filter that removes all characters outside the Unicode blocks corresponding to the target language. Stories were processed using language-specific regular expressions to preserve punctuation while eliminating Latin alphabet leakage.

\begin{table}[t]
\centering
\caption{Script normalization statistics after removing Latin characters and English tokens from stories.}
\label{tab:filter_stats}
\resizebox{\columnwidth}{!}{
\begin{tabular}{lccc}
\toprule
\textbf{Language} & \textbf{Stories} & \textbf{Modified} & \textbf{Modified (\%)} \\
\midrule
Assamese & 4,875 & 181 & 3.7 \\
Dogri & 4,924 & 102 & 2.1 \\
Konkani (GOM) & 4,879 & 123 & 2.5 \\
Gujarati & 12,856 & 1,767 & 13.7 \\
Kannada & 11,644 & 1,673 & 14.4 \\
Maithili & 4,872 & 130 & 2.7 \\
Malayalam & 11,216 & 3,103 & 27.7 \\
Manipuri & 4,870 & 218 & 4.5 \\
Nepali & 4,863 & 425 & 8.7 \\
Odia & 13,004 & 4,069 & 31.3 \\
Punjabi & 13,144 & 5,402 & 41.1 \\
Sanskrit & 4,873 & 257 & 5.3 \\
Santali & 4,883 & 120 & 2.4 \\
Sindhi & 4,881 & 606 & 12.4 \\
Tamil & 12,860 & 2,791 & 21.7 \\
Telugu & 10,924 & 3,029 & 27.7 \\
Urdu & 3,374 & 252 & 7.5 \\
\bottomrule
\end{tabular}}
\end{table}

\begin{table}[h]
\centering
\caption{Corpus statistics across the targeted Indic languages.}
\label{tab:dataset_stats}
\resizebox{\columnwidth}{!}{%
\begin{tabular}{llcc}
\toprule
\textbf{Code} & \textbf{Language} & \textbf{Stories} & \textbf{Tokens} \\
\midrule
\texttt{gu} & Gujarati & 12,856 & 9,858,511 \\
\texttt{kn} & Kannada & 11,644 & 9,890,334 \\
\texttt{ml} & Malayalam & 11,216 & 9,742,815 \\
\texttt{or} & Odia & 13,004 & 9,506,384 \\
\texttt{pa} & Punjabi & 13,144 & 9,669,977 \\
\texttt{ta} & Tamil & 12,860 & 9,840,128 \\
\texttt{te} & Telugu & 10,924 & 9,865,743 \\
\midrule
\texttt{as} & Assamese & 4,875 & 3,088,287 \\
\texttt{doi} & Dogri & 4,924 & 2,556,071 \\
\texttt{gom} & Konkani & 4,879 & 2,437,488 \\
\texttt{mai} & Maithili & 4,872 & 2,363,974 \\
\texttt{mni} & Manipuri & 4,870 & 71,024 \\
\texttt{ne} & Nepali & 4,863 & 2,309,707 \\
\texttt{sa} & Sanskrit & 4,873 & 2,605,271 \\
\texttt{sat} & Santali & 4,883 & 6,555,546 \\
\texttt{sd} & Sindhi & 4,881 & 2,029,536 \\
\texttt{ur} & Urdu & 3,374 & 1,519,067 \\
\bottomrule
\end{tabular}%
}
\end{table}

\subsection{Lexical Diversity Analysis}

To evaluate the narrative diversity of the generated corpus, we computed
\textit{Distinct-n} metrics \citep{li2016diversity} across each language split.
Distinct-n measures the proportion of unique $n$-grams relative to the total
number of generated $n$-grams, providing an estimate of lexical variation.
Higher values indicate greater diversity and reduced repetition.

In addition, we measured the average cosine similarity between sentence
embeddings of randomly sampled story pairs within each language split. Story
embeddings were computed using a multilingual sentence encoder, and lower
similarity values indicate greater semantic variability.

Table~\ref{tab:diversity} reports representative diversity statistics across
several languages in the dataset.

Languages generated natively using the Sarvam-M model (e.g., Kannada,
Malayalam, Tamil, and Telugu) exhibit the highest lexical diversity, with
Distinct-2 scores exceeding 0.65. In contrast, languages produced through the
machine translation expansion pipeline display slightly lower diversity,
likely due to normalization effects introduced by translation systems.

Overall, the dataset demonstrates substantial lexical and semantic variability,
with Distinct-2 values ranging from 0.25 to 0.69 and average embedding
similarities between 0.11 and 0.38. These results indicate that the
combinatorial prompt generation framework successfully produces diverse
narrative structures while avoiding large-scale repetition.

\section{Training Small Language Models}

The dataset is specifically designed for the pre-training of compact
transformer-based language models ranging from 10M to 200M parameters.
Due to the constrained vocabulary and short narrative format, the corpus
facilitates rapid convergence while preserving grammatical structure and
basic narrative reasoning.

A typical training configuration may involve a decoder-only transformer
trained with causal language modeling objectives. Preliminary experiments
suggest that models trained on synthetic narrative corpora exhibit improved
sentence completion, story continuation, and character consistency when
compared to models trained purely on heterogeneous web text.

\section{Dataset Release}

The Multilingual TinyStories dataset is released as a collection of JSONL
files, with one file per language split. Each record contains a unique
identifier, language code, and story text field. The dataset is distributed
under the CC-BY-4.0 license and is hosted on Hugging Face\footnote{\url{https://huggingface.co/datasets/deeponh/multilingual-tinystories}} 
for easy integration with existing machine learning pipelines.

To facilitate reproducibility, we additionally release the full prompt
generation templates and filtering scripts used in the dataset construction
pipeline\footnote{\url{https://github.com/deeps73/multilingual-tinystories}}.

\section{Conclusion}
We presented Multilingual TinyStories, a scalable and synthetically curated dataset comprising 17 Indian languages. By leveraging a combinatorial prompt engineering framework on Sarvam-M alongside a robust Google Translate API projection pipeline, we established a vast corpus of simple narratives. This resource aims to catalyze research into efficient language modeling and cross-lingual NLP for underrepresented demographics across the Indian subcontinent.

\section{Limitations}
Because the foundational stories are generated by a language model, there is a risk of subtle semantic hallucinations. Furthermore, as the expanded corpus relies on machine translation via the Google Translate API, the secondary 10 languages may exhibit translation artifacts or lack strict idiomatic nuance compared to native human speech. Consequently, the dataset should be used cautiously for downstream tasks requiring high factual accuracy or complex semantic reasoning.
\bibliography{custom}

\appendix

\section{Appendix A : Dataset Examples}
\label{app:dataset_examples}
The following examples present randomly sampled stories from the Multilingual TinyStories dataset across several Indic languages.

\begin{figure*}
    \centering
    \includegraphics[width=1\linewidth]{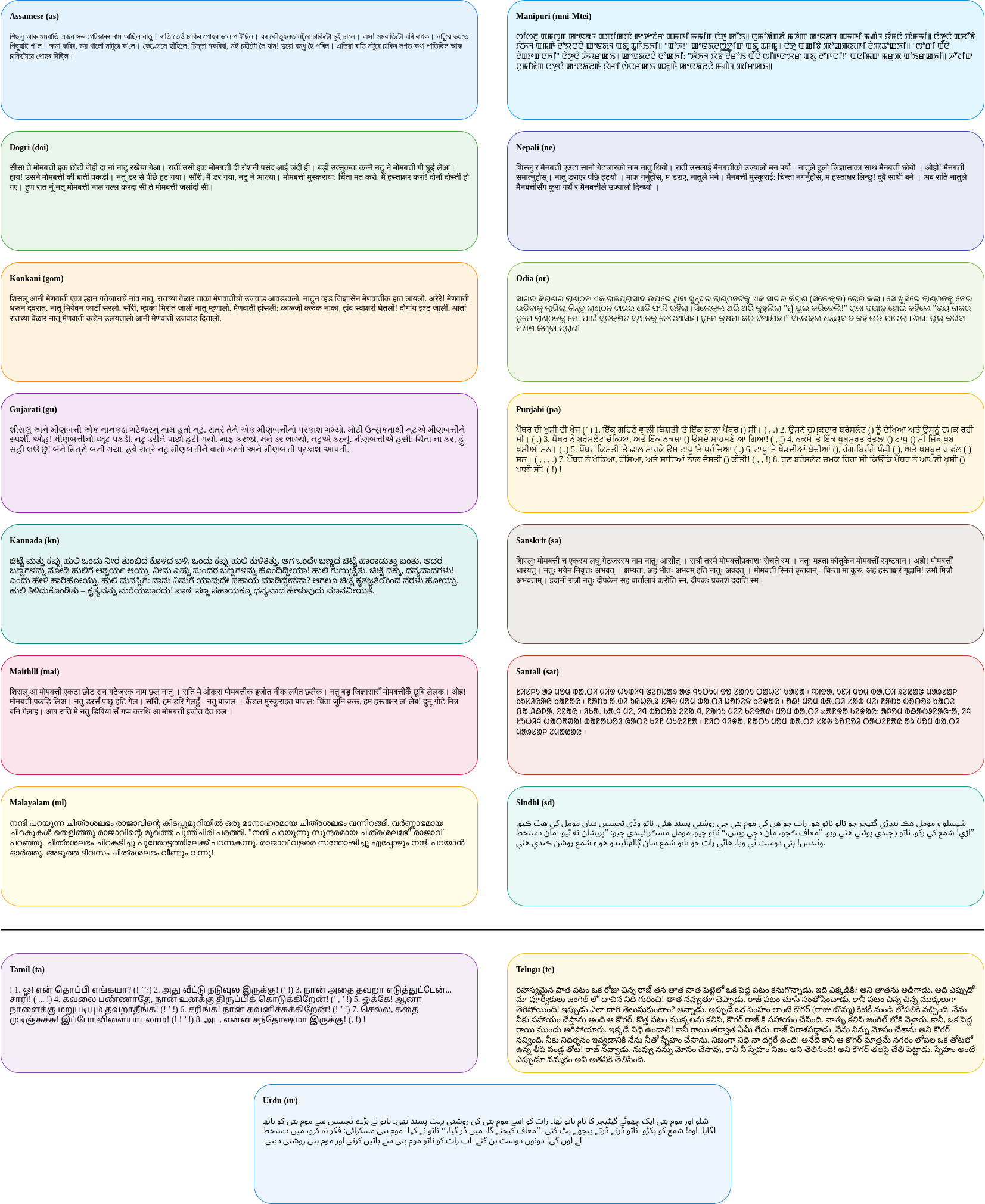}
    \caption{Samples from the Dataset}
    \label{fig:placeholder}
\end{figure*}

\section{Appendix B: Narrative Prompt Templates}
\label{app:prompt_templates}

The dataset generation pipeline employs a stochastic prompt generation
framework implemented in the \texttt{prompt\_generator} module. Rather
than using a single instruction format, the system samples from several
narrative prompt templates corresponding to different story archetypes.
Each template contains placeholders that are dynamically populated
during runtime with randomly sampled slot values.

The following examples illustrate representative prompt templates used
during generation.

\textbf{Problem-Solving Template}

\begin{quote}
Write a short children's story in \{Language\}. The story should include a
\{character\} in a \{setting\}. The character faces a problem involving a
\{object\}. The story should teach something about \{theme\}. Use simple
language suitable for a five-year-old child and write 5 to 8 sentences.
\end{quote}

\textbf{Adventure Template}

\begin{quote}
Write a children's adventure story in \{Language\}. A \{character\} is
exploring a \{setting\}. During the adventure they discover a
\{object\}. The story should teach a lesson about \{theme\}. Use simple
language and keep the story between 5 and 8 sentences.
\end{quote}

\textbf{Friendship Template}

\begin{quote}
Write a short children's story in \{Language\}. A \{character\} meets a
new friend in a \{setting\}. Together they try to solve a problem
involving a \{object\}. The story should teach something about
\{theme\}. Use simple language suitable for a young child.
\end{quote}

\textbf{Moral Lesson Template}

\begin{quote}
Write a short moral story in \{Language\}. The story takes place in a
\{setting\} and follows a \{character\}. The character faces a challenge
involving a \{object\} and learns something about \{theme\}. Keep the
language simple and limit the story to 5--8 sentences.
\end{quote}

\textbf{Mystery Template}

\begin{quote}
Write a short mystery story in \{Language\}. A \{character\} in a
\{setting\} discovers something unusual: a \{object\}. They try to
understand what happened. The story should teach a lesson about
\{theme\}. Write the story using simple language.
\end{quote}

These templates ensure that generated stories span multiple narrative
structures while maintaining a consistent level of linguistic simplicity.

\section{Appendix C: Prompt Slot Vocabulary}
\label{app:slot_vocab}

Narrative templates contain several variable slots that are populated
during prompt generation. These slots are filled by sampling from curated
vocabulary pools containing simple and culturally neutral concepts.

\textbf{Characters}

boy, girl, rabbit, elephant, robot, princess, farmer, bird,
kitten, turtle, fox, monkey, teacher

\textbf{Settings}

forest, village, riverbank, desert, mountain, jungle,
school, garden, beach, field

\textbf{Objects or Problems}

lost lantern, hidden treasure, broken bridge,
missing toy, locked door, magic seed,
lost book, strange map

\textbf{Themes}

honesty, kindness, curiosity, bravery,
friendship, patience, sharing, responsibility

\section{Appendix D: Prompt Combinatorial Space}
\label{app:prompt_space}
The prompt generation framework is designed to produce a large number
of unique narrative instructions through combinatorial slot sampling.
Each prompt template contains several variable slots that are filled
during runtime using randomly sampled vocabulary elements.

The primary slot categories used in the generation pipeline are:

\begin{itemize}
\item Characters ($N \approx 50$)
\item Settings ($N \approx 40$)
\item Objects or Problems ($N \approx 50$)
\item Themes or Moral Lessons ($N \approx 30$)
\end{itemize}

Each generated prompt is formed by sampling one element from each
vocabulary pool and inserting it into the corresponding template
slots. Under a simplified assumption that each slot is sampled
independently, the theoretical prompt space can be estimated as:

\[
50 \times 40 \times 50 \times 30 = 3{,}000{,}000
\]

This combinatorial design allows the pipeline to generate millions
of structurally distinct prompts while maintaining a consistent
instruction format. As a result, the language model is exposed to a
broad range of narrative configurations, reducing prompt repetition
and encouraging greater diversity in the generated stories.

\end{document}